\documentclass[lettersize,journal]{IEEEtran}
\usepackage{amsmath,amsfonts}
\usepackage{algorithmic}
\usepackage{algorithm}
\usepackage{array}
\usepackage[caption=false,font=normalsize,labelfont=sf,textfont=sf]{subfig}
\usepackage{textcomp}
\usepackage{stfloats}
\usepackage{url}
\usepackage{verbatim}
\usepackage{graphicx}
\usepackage{cite}
\usepackage{booktabs} 
\usepackage{multirow}
\usepackage[table]{xcolor}
\usepackage{siunitx}
\usepackage{float}
\usepackage{amsmath,amsfonts,amssymb}
\usepackage{capt-of}

\newcounter{subfig}

\let\oldtwocolumn\twocolumn
\renewcommand\twocolumn[1][]{%
  \oldtwocolumn[{#1}{%
    \begin{center}
      \refstepcounter{figure}

      \vspace{-8pt} 
      \includegraphics[width=0.98\textwidth]{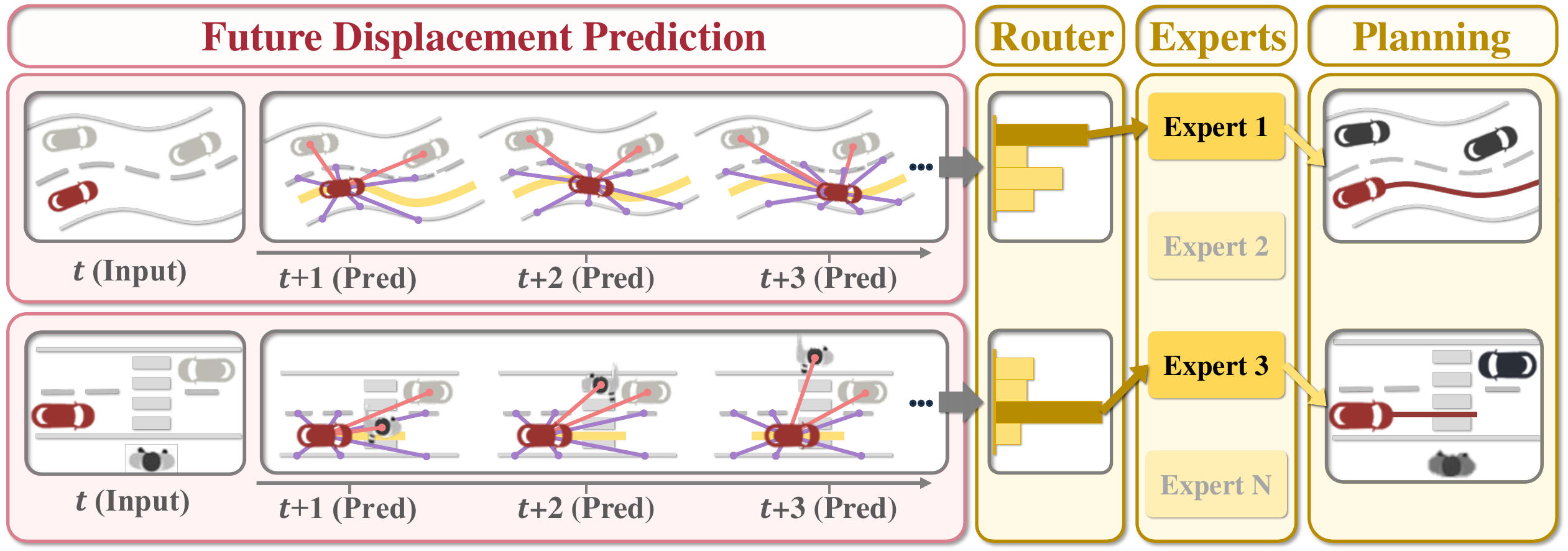}
      \vspace{-4pt}

      \addtocounter{figure}{-1}
      \captionof{figure}{\textbf{Key concept of CarPLAN.}
      CarPLAN predicts future displacement vectors between the AV and scene elements, enhancing awareness of relative spacing.
      To handle diverse dynamic scenes, CarPLAN adopts a mixture-of-experts (MoE) decoder where experts specialize in different contexts.
      The router dynamically selects experts conditioned on the current scene, enabling context-adaptive planning.}
      \label{fig:intro}

      \vspace{2pt}
    \end{center}
  }]%
}

\begin{document}

\title{CarPLAN: Context-Adaptive and Robust Planning with Dynamic Scene Awareness for Autonomous Driving}

\author{
    Junyong Yun$^{1*}$, Jungho Kim$^{2*}$, ByungHyun Lee$^1$, Dongyoung Lee$^3$, Sehwan Choi$^1$, Seunghyeop Nam$^2$, \\
    Kichun Jo$^4$, and Jun Won Choi$^3$ \\
    \thanks{$^1$J. Yun, B. Lee, and S. Choi are with the Department of Artificial Intelligence, Hanyang University, 04763, Republic of Korea. (e-mail:  jyyun@spa.hanyang.ac.kr, bhlee@spa.hanyang.ac.kr, sehwanchoi@spa.hanyang.ac.kr)}
    \thanks{$^2$J. Kim and S. Nam are with the Interdisciplinary Program in Artificial Intelligence, Seoul National University, 08826, Seoul, Republic of Korea (e-mail: jhkim@spa.snu.ac.kr, shnam@adr.ac.kr)}
    \thanks{$^3$D. Lee and J. W. Choi are with the Department of Electrical and Computer Engineering, Seoul National University, 08826, Seoul,  Republic of Korea. (e-mail: dylee@spa.snu.ac.kr,   junwchoi@snu.ac.kr)}
    \thanks{$^4$K. Jo is with the Department of Automotive Engineering, Hanyang University, 04763, Republic of Korea. (e-mail:  kichunjo@hanyang.ac.kr)}
    \thanks{\textit{(Corresponding author: Jun Won Choi)}}
    \thanks{$^*$denotes equal contribution.}
}

\markboth{}{}


\maketitle


\begin{abstract}
Imitation learning (IL) is widely used for motion planning in autonomous driving due to its data efficiency and access to real-world driving data. For safe and robust real-world driving, IL-based planning requires capturing the complex driving contexts inherent in real-world data and enabling context-adaptive decision-making, rather than relying solely on expert trajectory imitation. In this paper, we propose CarPLAN, a novel IL-based motion planning framework that explicitly enhances driving context understanding and enables adaptive planning across diverse traffic scenarios. Our contributions are twofold:
We introduce Displacement-Aware Predictive Encoding (DPE) to improve the model’s spatial awareness by predicting future displacement vectors between the Autonomous Vehicle (AV) and surrounding scene elements. This allows the planner to account for relational spacing when generating trajectories. In addition to the standard imitation loss, we incorporate an augmented loss term that captures displacement prediction errors, ensuring planning decisions consider relative distances from other agents.
To improve the model’s ability to handle diverse driving contexts, we propose Context-Adaptive Multi-Expert Decoder (CMD), which leverages the Mixture of Experts (MoE) framework. CMD dynamically selects the most suitable expert decoders based on scene structure at each Transformer layer, enabling adaptive and context-aware planning in dynamic environments. We evaluate CarPLAN on the nuPlan benchmark and demonstrate state-of-the-art performance across all closed-loop simulation metrics. In particular, CarPLAN exhibits robust performance on challenging scenarios such as Test14-Hard, validating its effectiveness in complex driving conditions. Additional experiments on the Waymax benchmark further demonstrate its generalization capability across different benchmark settings.
\end{abstract}


\section{Introduction}


\IEEEPARstart{A}{utonomous} driving requires robust motion planning to navigate complex and dynamic traffic environments safely and efficiently.
Traditional planning approaches have predominantly relied on rule-based algorithms \cite{IDM, PDM, rule_1, thrun2006stanley, bacha2008odin, leonard2008perception, urmson2008autonomous, chen2015deepdriving}, where the behavior of an autonomous vehicle (AV) is explicitly defined through handcrafted rules.
While these methods provide interpretable decision-making, they demand substantial engineering effort to design and modify rules in various driving environments, resulting in a lack of adaptability to new traffic scenarios. 
To overcome these challenges, Imitation Learning (IL) \cite{IL_1, IL_2, IL_3, IL_4} has emerged as a promising alternative, enabling models to learn driving policies directly from expert human demonstrations.
By leveraging large-scale real-world datasets, such as nuPlan \cite{nuplan}, IL-based approaches can capture a wide range of driving behaviors without relying on manually defined rules or reward functions.

\begin{figure}[t]
    \centering
    \vspace{-8pt} 
    \includegraphics[width=\linewidth]{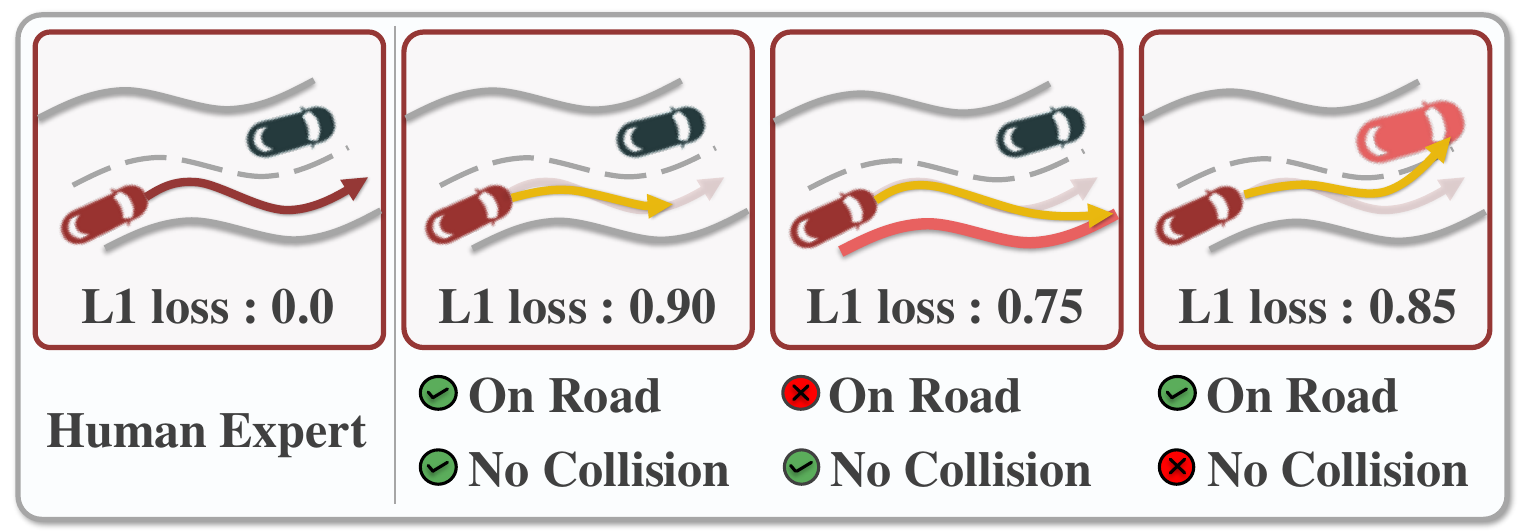}
    \caption{\textbf{Challenges in imitation learning.} 
        Minimizing imitation loss alone does not guarantee safe planning. 
        This example demonstrates that a lower L1 loss could result in lane departures or collisions.}
    \label{fig:intro1}
\end{figure}

Most IL-based planners are trained by minimizing trajectory imitation losses that measure the distance between predicted trajectories and ground-truth (GT) trajectories.
However, such objectives are not always aligned with real-world driving requirements, such as collision avoidance or road boundary compliance.
In some cases, IL models prioritize generating trajectories that closely match the GT rather than actively avoiding collisions, as illustrated in Figure \ref{fig:intro1}.
This observation indicates that motion planning is not merely about matching human trajectories, but about correctly understanding the driving context and selecting appropriate actions accordingly.
In real-world driving situations, diverse environmental factors (e.g., vehicles, pedestrians, and road structures) interact simultaneously in complex ways, and different combinations of these factors lead to distinct decision-making criteria.
Therefore, robust IL-based planning requires both accurate context understanding of the driving scene and the ability to perform context-adaptive planning, adjusting driving strategies according to the situation.

Accurately understanding driving context requires the ability to capture how the surrounding environment evolves over time.
This evolution is not solely characterized by the independent motions of nearby objects, but by the evolving spatial relationships between the autonomous vehicle and its surrounding environment.
Such relational dynamics reflect driving patterns implicitly inherent in human demonstration data and should be explicitly learned for robust IL-based planning.
However, many existing IL-based planners \cite{occplanner, gameformer, ConvLSTMplanner, rasc, PLUTO, BeTopNet, Diffusion-Planner, carplanner} predominantly rely on independent future predictions of individual agents, often limited to vehicles, without sufficiently modeling their spatial relationships with broader scene elements such as road structures or static obstacles, leading to insufficient relational modeling in scene representations.

Beyond context understanding, context-adaptive planning requires the ability to adjust driving strategies across situations.
Even when driving situations appear similar, appropriate driving strategies can differ depending on the surrounding context and interaction patterns, making them difficult to handle with a single fixed policy.
However, existing IL-based planners rely on a single shared policy network, which biases learning toward common scenarios and limits the model's ability to reflect behavior variations required across different contexts.
As a result, these planners are vulnerable to adopting adaptive driving strategies in complex or rarely encountered driving situations.


To address these limitations, we propose CarPLAN (Context-Adaptive and Robust Planner), a novel IL-based motion planning framework designed to jointly achieve nuanced context understanding and context-adaptive planning.
CarPLAN is built on the key insight that robust planning requires not only capturing rich contextual information from driving scenes, but also adapting driving strategies based on the captured context.
Accordingly, CarPLAN explicitly models the contextual structure of the scene and leverages it to guide adaptive decision-making across diverse driving situations.

Specifically, we aim to equip the encoder with enhanced spatial awareness of surrounding elements—such as vehicles, pedestrians, lanes, and road boundaries—relative to the ego vehicle.
To this end, we introduce the Displacement-Aware Predictive Loss, a self-supervised objective that enforces predictive constraints on the displacement between the ego vehicle and surrounding scene elements.
We design the Displacement-Aware Predictive Encoder (DPE), which forecasts future displacement vectors between the AV and nearby entities based on their current relative positions (see Figure~\ref{fig:intro}).
The Displacement-Aware Predictive Loss measures the deviation between the predicted and ground-truth displacement vectors, guiding the encoder to encode spatial relationships with surrounding objects in its feature representations.
Importantly, the DPE functions solely as a training-time supervisory signal and can be deactivated during inference, introducing no additional runtime computational overhead.

We also introduce the Context-Adaptive Multi-Expert Decoder (CMD), which leverages a set of specialized expert networks tailored to different driving contexts.
We integrate the Mixture of Experts (MoE) framework, a proven approach in large language models for enhancing performance on complex tasks \cite{switchtransformer, deepseekmoe, Gshared, moe_llm_1, moe_llm_2}, into the decoder in our CarPLAN.
By dynamically selecting the most relevant experts based on the scene context, our approach improves adaptability and robustness in decision-making across diverse driving conditions.
To support this mechanism, we further design a Scene-Aware Router, which analyzes the dynamic scene structure to determine the most suitable expert network for each planning instance.

We evaluate CarPLAN on the widely used nuPlan benchmark \cite{nuplan}.
Our CarPLAN achieved state-of-the-art (SOTA) performance across all closed-loop simulation metrics.
In particular, it attained a remarkable score of 84.6 in the reactive simulation and 91.4 in the non-reactive simulation on the Val14 benchmark.
Furthermore, CarPLAN demonstrated robust performance on the more challenging Test14-Hard, which includes complex and difficult driving scenarios.
To further examine generalization ability, we also conducted a reactive simulation on the Waymax benchmark \cite{waymax}, where CarPLAN consistently achieved performance improvements over baseline methods across diverse scenarios.


The main contributions of this paper are summarized below:
\begin{itemize}
    \item 
  We propose CarPLAN, a novel approach for IL-based motion planning that enhances the model's awareness of its surroundings, the proposed DPE predicts future displacement vectors between AV and objects in the scene.
  
  \item CarPLAN adopts MoE for context-adaptive planning in diverse environments. This idea enhances the planner's ability to interpret and adapt to varying scenes, resulting in improved trajectory planning. 
   
    \item CarPLAN achieves SOTA performance on nuPlan benchmark and demonstrates strong robustness and generalization on challenging benchmarks, including Test14-Hard and Waymax.

\end{itemize}

\section{Related works}
\label{sec:related}

\subsection{Imitation Learning-based Planning}
Imitation Learning \cite{codevilla2019exploring, dauner2023parting, ross2011reduction, wen2020fighting, phong2023truly, zhai2023rethinking} has emerged as a prominent approach for motion planning in autonomous driving, as it enables models to learn directly from expert demonstrations without relying on handcrafted rules or explicit reward functions.
PlanTF \cite{PlanTF} addressed spurious correlations in imitation learning by using only the AV’s present state and introduced the AV states dropout that was selectively masked during training to prevent reliance on specific kinematic states.
PLUTO \cite{PLUTO} mitigated causal confusion in imitation learning through contrastive learning with data augmentation, using negative samples such as leading agent dropout and traffic light inversion.
RaSc \cite{rasc} introduced a risk-aware planning approach by leveraging pairwise TTC with other agents. To learn risk awareness, it enforced self-consistency by predicting pairwise TTC for all agents and aligning its planned trajectories with these predictions.
BeTopNet \cite{BeTopNet} enhanced prediction and planning consistency by explicitly modeling multi-agent future behaviors through topological reasoning based on braid theory.
Diffusion-Planner \cite{Diffusion-Planner} proposed a transformer-based diffusion model that refines future trajectories through iterative denoising, capturing interaction-aware behaviors and enabling rule-free planning via classifier guidance.
CarPlanner \cite{carplanner} introduced a consistent auto-regressive trajectory planner that sequentially generates future steps conditioned on previous outputs, incorporating mode-consistent sampling and expert-guided rewards.

\subsection{Mixture of Experts}
The Mixture of Experts (MoE) architecture \cite{jacobs1991adaptive} dynamically activates multiple specialized experts based on input contexts, efficiently allocating model capacity to enhance task performance. Recent advances have demonstrated its effectiveness across diverse domains \cite{Gshared,MoE_nlp1,MoE_nlp2,MoE_nlp3,MoE_nlp4,MoE_nlp5,MoE_cv1,MoE_cv2,MoE_cv3,MoE_cv4,MoE_cv5,MoE_cv6}. Switch Transformer \cite{switchtransformer} utilized sparse expert activation for computational efficiency and large-scale model scalability. DeepseekMoE \cite{deepseekmoe} introduced shared and routed experts to minimize redundancy by clearly separating common and specialized knowledge. Expert-choice routing \cite{expertchoice} allowed each expert to specialize in specific subsets of input distributions, increasing overall model adaptability. FLAN-MOE \cite{FLAN_MoE} demonstrated improved performance by integrating instruction-tuning with the MoE framework, enhancing experts' task-specific specialization.

\begin{figure*}[!t]
    \centering
    \includegraphics[scale=0.58]{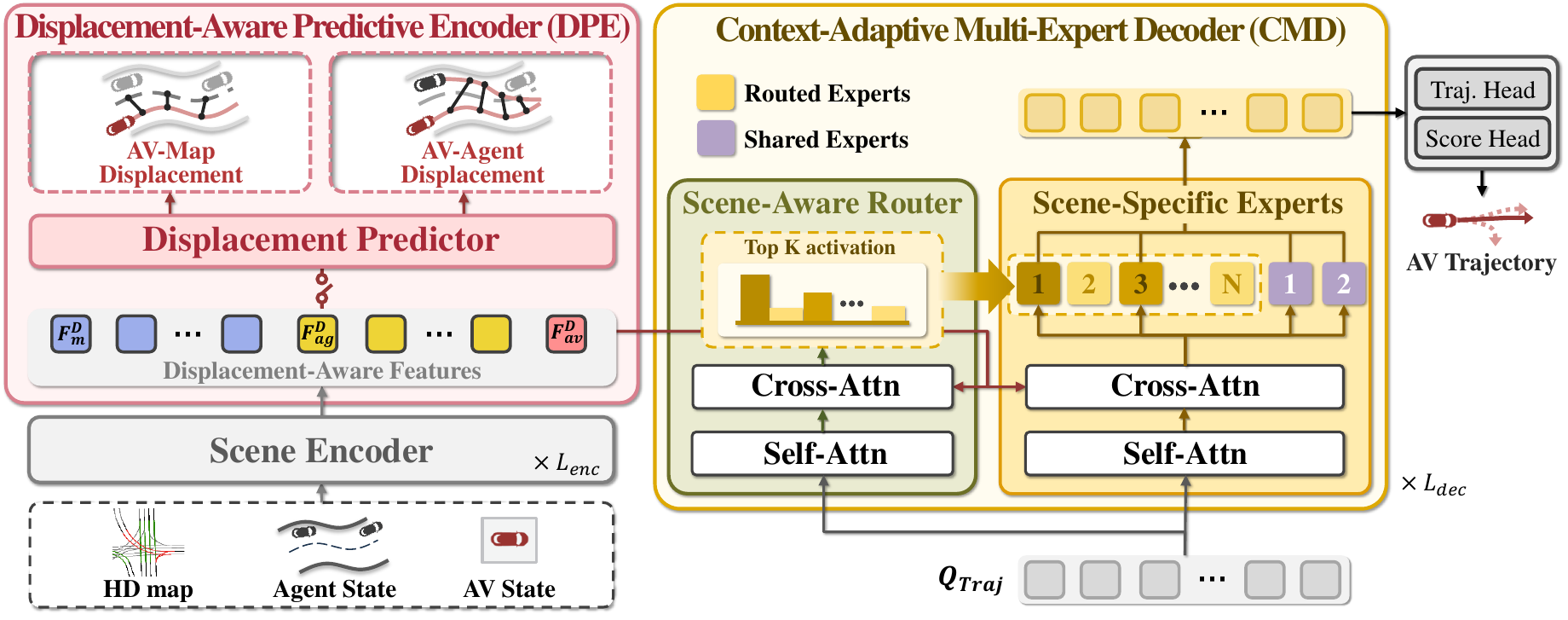}
    \caption{\textbf{Overall structure  of CarPLAN}. 
CarPLAN comprises two main networks: DPE and CMD. DPE is trained to predict the displacements or velocity between the AV and surrounding scene elements at each future timestep, generating Displacement-Aware Features. CMD utilizes multiple experts, dynamically selected by the Scene-Aware Router, to generate the AV's future trajectory.
    }
    \label{architecture_fig}
\end{figure*}

\section{CarPLAN}
\subsection{Overview}
The overall architecture of CarPLAN is depicted in Figure \ref{architecture_fig}. CarPLAN first encodes input data consisting of the AV, surrounding agents, and high-definition (HD) map through Scene Encoder, generating comprehensive scene context features. Subsequently, the Displacement Predictor predicts future displacement between the AV and surrounding scene elements, producing Displacement-Aware Features. These features are trained to encode relative spacing from surrounding objects through the Displacement-Aware Predictive Loss, which enforces predictive self-supervised constraints on displacement vectors. Next, given a Trajectory Query \(Q_{traj}\), the Scene-Aware Router dynamically selects the most relevant expert networks by analyzing the \(Q_{traj}\) jointly with the Displacement-Aware Features. Specifically, it selects the top-$K$ Routed Experts. The selected experts extract scenario-specific features tailored to the current driving context, while Shared Experts extract general driving features. Both features are aggregated and then used by the Trajectory Head and Score Head to generate the AV's $M$ multi-modal future trajectories with their confidence scores.

\subsection{Problem Formulation}

This work addresses the critical challenge of AV planning in complex urban environments, aiming to generate safe future trajectories that consider the scene contexts and interactions with dynamic agents (e.g.,
vehicles, bicycles, and pedestrians). We consider a driving scene consisting of an AV, surrounding agents, and a vectorized map. The state information of all agents over the historical horizon \( T_h \) is represented as  $
A_{0:N_a} = \{A_a^{-T_h:0}\}_{a=0}^{N_a}
$, where \( A_0 \) corresponds to the states of the AV, and \( A_{1:N_a} \) represents those of the surrounding agents.
Each state includes the position, heading, velocity, size, and category of the agent. For the vectorized map, we define \( P_{1:N_p} \) as a collection of map elements, represented as points that compose lane geometry, crosswalks, road boundaries, and so on.
Additionally, multiple centerlines in the form of polylines are provided as guidance for the AV to navigate toward its goal position, denoted as \( L_g = \{L_k\}_{k=1}^{N_K} \), where \( N_K \) represents the number of centerlines connected to the goal in the current scene. The goal position corresponds to the last point of a long logged sequence, which is much longer than the planned trajectory's prediction horizon.
Given all these conditions, the objective of planning is to generate the $M$ trajectory states of AV over \(T_f\) future time steps, $\hat{Y} = \{\hat{y}_m^{1:T_f} \}_{m=1}^{M}$ and their corresponding confidence scores  $\hat{C} =\{\hat{c}_m\}_{m=1}^{M}$.

\subsection{Displacement-Aware Predictive Encoder}
DPE consists of the Scene Encoder and the Displacement Predictor. The Scene Encoder employs the Transformer Encoder \cite{multi-head-attention} to generate a unified representation of the overall scene, capturing interactions among traffic participants. The Displacement Predictor utilizes the unified scene representation to forecast the future displacement of the AV relative to each surrounding element.

\subsubsection{Scene Encoder}
The Scene Encoder processes \(A_{0}\), \(A_{1:N_a}\), and \(P_{1:N_p}\) to encode them jointly through Transformer.
Specifically, these elements are encoded into \(S_{av}\), \(S_{agent}\), and \(S_{map}\) using modality-specific encoding modules, as suggested in \cite{forecast-mae}.
These encoded features are concatenated as \(S = [S_{av}; S_{agent}; S_{map}]\), and then fed to a Transformer Encoder to generate the scene context features \(S' = [S'_{av}; S'_{agent}; S'_{map}] \in \mathbb{R}^{(1+N_a+N_p) \times D}\), where \([\cdot]\) denotes a concatenation operation.

\subsubsection{Displacement Predictor} The Displacement Predictor guides the Scene Encoder to generate Displacement-Aware Features \(F^{D}  = S'\), which encode information on relative spacing to scene elements. This is achieved by training DPE to predict future displacement vectors between the AV and surrounding traffic and dynamic objects.
To effectively capture the relational information between the AV and the neighboring agents, we concatenate the AV feature \(F^{D}_{av}\) with the agent features \(F^{D}_{agent}\) channel-wise, resulting in the features \(F_{agent}\). Similarly, we concatenate  \(F^{D}_{av}\) with the map features \(F^{D}_{map}\) channel-wise to obtain \(F_{map}\).  Then, after concatenating \(F_{agent}\) and \(F_{map}\), \(\text{DispHead}\) is applied  to predict the displacement vectors at \(T_f\) future timesteps, i.e., 
\begin{align}
    \hat{D}_{1:N_a+N_p}^{1:T_f} &= \text{DispHead}([F_{agent};F_{map}]),
\end{align}
where \(\text{DispHead}(\cdot)\) consists of two MLP layers.  Note that the process of predicting the future displacement is performed only during the training phase, not during inference.

 \subsection{Context-Adaptive Multi-Expert Decoder}
CMD utilizes the MoE framework \cite{deepseekmoe} to enable context-adaptive planning across diverse driving scenarios. It consists of multiple iterative decoding layers, each incorporating a Scene-Aware Router and Scene-Specific Experts.  
The Trajectory Query \( Q_{traj} \) is initialized from centerlines \(L_g\) using the encoding method proposed in \cite{PLUTO} and is iteratively refined across multiple layers.  

Scene-Specific Experts employ multiple expert decoders to generate the AV’s future trajectory based on Displacement-Aware Features and Trajectory Queries. We utilize two types of experts: \( N \) Routed Experts and \( N_s \) Shared Experts. These expert decoders refine the Trajectory Query while capturing semantic scene information.  

The Scene-Aware Router processes Displacement-Aware Features to extract high-level scene representations, which are then used to select the most suitable Top-$K$ Routed Experts. Notably, Shared Experts are always activated, regardless of expert selection. Finally, the Trajectory Head and Score Head predict the AV’s future trajectories and their corresponding confidence scores.

\subsubsection{Scene-Aware Router}
The Scene-Aware Router produces the probability scores for Routed Experts based on Trajectory Query \(Q_{traj}\) and Displacement-Aware Features \(F^{D}\). 
Using \( Q_{traj} \) as the query and the Displacement-Aware Features \( F^{D} \) as the key and value, the Scene-Aware Router performs
\begin{align}
Q_{Routed} = \text{CA}(\text{SA}(Q_{traj}), F^{D}, F^{D}),
\label{eq1}
\end{align}
where \( \text{SA}(\cdot) \) and \( \text{CA}(\cdot) \) represent self-attention and cross-attention operations, respectively \cite{multi-head-attention}.

Therefore, \( Q_{Routed} \) captures the AV's driving intention as well as its relationship with the surrounding scene. Subsequently, \( Q_{Routed} \) is passed through an MLP layer followed by a softmax function to predict the scores for the \( N \) Routed Experts. For each query, we select the top-\( K \) Specific Experts based on these scores:
\begin{align}
\{R_{i}\}_{i\in E} = \text{Top-$K$}(\text{SoftMax}(\text{MLP}(Q_{Routed}))),
\end{align}
where \( E \) represents the indices of the experts selected by the Top-$K$ operation, and \( R_i \) denotes the score of the \( i \)-th expert selected for each query.

\subsubsection{Scene-Specific Experts}
Scene-Specific Experts transform the Trajectory Query \( Q_{traj} \) while integrating relevant information for AV planning.  
Before this transformation, an operation similar to Equation \ref{eq1} is performed using \( Q_{traj} \) as the query and the Displacement-Aware Features \( F^{D} \) as the key and value, producing the features \( Q'_{Expert} \).  
Next, the top-$K$ Routed Experts and Shared Experts, each composed of Feed-Forward Networks (FFNs), independently process each element of \( Q'_{Expert} \). Note that the top-$K$ Routed Experts are selected by the Scene-Aware Router.  

Finally, the features produced by the Routed Experts and Shared Experts are combined to generate the scene-aware feature \( F_{sa} \), formulated as  
\begin{equation}
    F_{sa} = \sum_{i=1}^{N_s}S_i(Q'_{Expert}) + \sum_{i \in E} R_i \cdot E_i(Q'_{Expert}),
\end{equation}
where \( S_i(\cdot) \) and \( E_i(\cdot) \) denote the outputs of the \( i \)-th Shared Expert and the \( i \)-th Routed Expert, respectively. Notice that the outputs of the top-$K$ Routed Experts are weighted by the scores produced by the Scene-Aware Router.   

After passing through a total of \( L_{dec} \) layers, the extracted final scene-aware features \( F_{sa} \) are used to generate the AV's future trajectory \( \hat{Y} \) and confidence score \( \hat{C} \) through a regression head and a classification head.

\subsection{Learning Process}
The total loss $L_{total}$ used to train CarPLAN is given by
\begin{equation}
    L_{total} = L_{plan} + L_{disp} + L_{bal},
\end{equation}
where $L_{plan}$, $L_{disp}$, and $L_{bal}$ represent the Planning Loss, Displacement-Aware Predictive Loss, and Expert Balancing Loss, respectively. The Planning Loss $L_{plan}$ is computed using the smooth L1 loss and cross-entropy loss, $L_{plan} = L1_{smooth}(Y, \hat{Y}) + \text{CrossEntropy}(C, \hat{C})$,
where $Y$ and $C$ are the AV's ground truth trajectory and score. The Displacement-Aware Predictive Loss $L_{disp}$ is calculated using the smooth L1 loss from
\begin{equation}
\begin{aligned}
 L_{disp} &= L1_{smooth}(D_{1:N_a+N_p}^{1:T_f}, \hat{D}_{1:N_a+N_p}^{1:T_f}) \\
    D_{1:N_a+N_p}^{1:T_f} &= \{(x_a^1 - x_0^1),..., (x_a^{T_f} - x_0^{T_f})\}_{a=1:N_a+N_p},
 \end{aligned}
\end{equation}
where $x_{a}^t$ denotes the positions of agent $a$ at $t$ timestep and $D_{1:N_a+N_p}^{1:T_f}$ is the ground truth displacement of surrounding scene elements over $T_f$ horizon. Lastly, we employ the Expert Balance Loss $L_{bal}$ to guarantee a balanced selection of experts \cite{MoE_nlp1}.

\newcommand{\B}[1]{{\bfseries #1}}
\newcommand{\U}[1]{{\underline{#1}}}
\newcommand{\BU}[1]{{\bfseries\underline{#1}}}

\begin{table*}[t]
\centering
\caption{Closed-loop simulation results on Val14, Test14-Hard, and Test14-Random. The ``--'' symbol means the metric is unknown.}
\label{tab:closed_loop_main}
\setlength{\tabcolsep}{4.2pt}
\renewcommand{\arraystretch}{1.10}

\begin{tabular}{
l l l
S[table-format=2.1] S[table-format=2.1] S[table-format=2.1]
S[table-format=2.1] S[table-format=2.1] S[table-format=2.1]
S[table-format=2.1] S[table-format=2.1] S[table-format=2.1]
}
\toprule
\multirow{2}{*}{\textbf{Type}} & \multirow{2}{*}{\textbf{Planner}} & \multirow{2}{*}{\textbf{Venue}} &
\multicolumn{3}{c}{\textbf{Val14}} &
\multicolumn{3}{c}{\textbf{Test14-Hard}} &
\multicolumn{3}{c}{\textbf{Test14-Random}} \\
\cmidrule(lr){4-6}\cmidrule(lr){7-9}\cmidrule(lr){10-12}
& & & {\textbf{OLS}} & {\textbf{CLS-NR}} & {\textbf{CLS-R}}
     & {\textbf{OLS}} & {\textbf{CLS-NR}} & {\textbf{CLS-R}}
     & {\textbf{OLS}} & {\textbf{CLS-NR}} & {\textbf{CLS-R}} \\
\midrule

\multirow{8}{*}{Learning} &
GameFormer \cite{gameformer} & ICCV'23 &
70.3 & 13.3 & \multicolumn{1}{c}{--} &
61.7 & 7.1 & 6.7 &
68.4 & 11.4 & 9.3 \\

& PDM-Open \cite{PDM} & CoRL'23 &
85.9 & 50.2 & 54.8 &
79.1 & 33.5 & 35.8 &
84.1 & 52.8 & 57.2 \\

& UrbanDriver \cite{urbandriver} & PMLR'22 &
84.2 & 68.6 & 64.8 &
77.2 & 50.4 & 50.0 &
81.2 & 51.8 & 67.2 \\

& PlanTF \cite{PlanTF} & ICRA'24 &
88.7 & 85.3 & 77.1 &
\U{84.9} & 69.7 & 61.6 &
87.8 & 85.6 & 79.6 \\

& PLUTO \cite{PLUTO} & arXiv'24 &
\U{89.3} & 88.9 & 80.0 &
81.4 & 74.0 & 59.7 &
87.8 & \U{89.9} & 78.6 \\

& RaSc \cite{rasc} & ECCV'24 &
87.0 & 86.0 & 75.0 &
\B{86.0} & 75.0 & 63.0 &
\B{91.0} & 86.0 & 76.0 \\

& BeTopNet \cite{BeTopNet} & NeurIPS'24 &
\multicolumn{1}{c}{--} & 88.3 & \U{83.7} &
84.0 & \U{77.1} & 68.8 &
87.6 & \B{90.2} & \U{85.7} \\

& Diffusion-Planner \cite{Diffusion-Planner} & ICLR'25 &
76.2 & \U{89.9} & 82.8 &
66.2 & 76.0 & \U{69.2} &
73.0 & 89.2 & 82.9 \\

\addlinespace[1pt]
\rowcolor[gray]{0.90}
& \B{CarPLAN (Ours)} & \multicolumn{1}{c}{--} &
\B{89.5} & \B{91.4} & \B{84.6} &
81.4 & \B{78.9} & \B{72.5} &
\U{88.4} & \B{90.2} & \B{87.1} \\
\bottomrule
\end{tabular}
\end{table*}

\section{Experiments}
\label{sec:experiments}

\begin{table}[t]
    \centering
    \renewcommand{\arraystretch}{1.2}
    \setlength{\tabcolsep}{2pt}
    \fontsize{8pt}{8pt}\selectfont
    \caption{Performance on the nuPlan benchmark with post-processing.}
    \label{table:hybrid}

    \begin{tabular}{l l  c c  c c  c c}
        \toprule[1.2pt]
        \multirow{2}{*}{\textbf{Type}} &
        \multirow{2}{*}{\textbf{Planner}} &
        \multicolumn{2}{c}{\textbf{Val14}} &
        \multicolumn{2}{c}{\textbf{Test14-Hard}} &
        \multicolumn{2}{c}{\textbf{Test14-Random}} \\
        \cmidrule(lr){3-4}
        \cmidrule(lr){5-6}
        \cmidrule(lr){7-8}
        & 
        & \textbf{NR} & \hspace{4pt}\textbf{R}
        & \hspace{7pt}\textbf{NR} & \textbf{R}
        & \hspace{9pt}\textbf{NR} & \textbf{R} \\
        \midrule

        \textcolor[gray]{0.5}{Expert} &
        \textcolor[gray]{0.5}{Log-replay} &
        \textcolor[gray]{0.5}{93.5} & \hspace{4pt}\textcolor[gray]{0.5}{75.9} &
        \hspace{7pt}\textcolor[gray]{0.5}{86.0} & \hspace{4pt}\textcolor[gray]{0.5}{65.8} &
        \hspace{9pt}\textcolor[gray]{0.5}{94.0} & \textcolor[gray]{0.5}{75.9} \\
        \midrule[0.1pt]

        \multirow{2}{*}{Rule} &
        IDM \cite{IDM} &
        75.6 & \hspace{4pt}77.3 &
        \hspace{7pt}56.2 & \hspace{4pt}62.3 &
        \hspace{9pt}70.4 & 74.4 \\
        &
        PDM-Closed \cite{PDM} & 
        92.8 & \hspace{4pt}92.1 &
        \hspace{7pt}66.0 & \hspace{4pt}76.1 &
        \hspace{9pt}90.2 & 91.2 \\
        \midrule[0.1pt]

        \multirow{7}{*}{Learning} &
        GameFormer \cite{gameformer} &
        82.9 & \hspace{4pt}83.8 &
        \hspace{7pt}68.7 & \hspace{4pt}67.1 &
        \hspace{9pt}83.9 & 82.1 \\
        &
        PDM-Hybrid \cite{PDM} &
        92.8 & \hspace{4pt}92.1 &
        \hspace{7pt}66.0 & \hspace{4pt}76.1 &
        \hspace{9pt}90.2 & 91.2 \\
        &
        RaSc \cite{rasc} &
        91.0 & \hspace{4pt}86.0 &
        \hspace{7pt}77.0 & \hspace{4pt}65.0 &
        \hspace{9pt}91.0 & 82.0 \\
        &
        PLUTO \cite{PLUTO} &
        93.2 & \hspace{4pt}76.8 &
        \hspace{7pt}\underline{80.1} & \hspace{4pt}76.9 &
        \hspace{9pt}92.2 & 90.2 \\
        &
        CarPlanner \cite{carplanner} &
        - & \hspace{4pt}- &
        \hspace{7pt}- & \hspace{4pt}- &
        \hspace{9pt}94.1 & 91.1 \\
        &
        Diffusion-Planner \cite{Diffusion-Planner} &
        \underline{94.3} & \hspace{4pt}\underline{92.9} &
        \hspace{7pt}78.9 & \hspace{4pt}\textbf{82.0} &
        \hspace{9pt}\textbf{94.8} & \underline{91.7} \\

        \rowcolor[gray]{0.90}
        & \textbf{CarPLAN (Ours)} &
        \textbf{95.0} & \hspace{4pt}\textbf{93.8} &
        \hspace{7pt}\textbf{82.2} & \hspace{4pt}\textbf{82.0} &
        \hspace{9pt}\underline{94.6} & \textbf{92.5} \\
        \bottomrule[1.2pt]
    \end{tabular}

    \vspace{-8pt}
\end{table}

\subsection{Experimental Settings}
\subsubsection{Benchmark and Metrics}
We conducted extensive evaluations on the nuPlan \cite{nuplan} dataset, which contains 1,300 hours of real-world urban driving data spanning 75 scenario types.  
The Val14 \cite{PLUTO}, Test14-Hard, and Test14-Random \cite{PlanTF} benchmarks are used to assess performance across both general and challenging driving scenarios, with evaluation metrics based on the Open-Loop Score (OLS) and Closed-Loop Score (CLS).  
OLS is computed using distance errors between the planned trajectories and the logged trajectories. In contrast, CLS assesses the outcomes of driving simulations generated from the model’s predictions, considering factors such as safety, efficiency, and ride comfort. CLS is further divided into CLS-NR (Non-Reactive) and CLS-R (Reactive), depending on how surrounding agents are simulated.  
In CLS-NR, agents strictly follow the logged trajectories from the dataset, reflecting real-world driving conditions. In CLS-R, agents adapt their behavior based on the Intelligent Driver Model (IDM) \cite{IDM}, allowing them to react dynamically to the ego vehicle.

In addition to the nuPlan benchmark \cite{nuplan}, we further evaluated the proposed model on the Waymax benchmark \cite{waymax} to assess its generalization capability across benchmarks. Waymax is a large-scale reactive closed-loop simulation framework built on the Waymo Open Motion Dataset (WOMD) \cite{womd}, providing real-world driving environments with diverse traffic interactions. Similar to the CLS-R setting in nuPlan, the ego vehicle is controlled by the planner, while surrounding agents react dynamically within the simulation.
Closed-loop performance in Waymax is evaluated using Arrival Rate (AR), Off-road Rate (OR), Collision Rate (CR), and Progress Rate (PR). AR measures whether the ego vehicle successfully completes the driving task without safety violations. OR and CR capture off-road and collision events, respectively, while PR reflects progress along the expert driving route.

\noindent
\subsubsection{Implementation Details}
We utilized agent states from the past \( T_h = 2 \) seconds, sampled at 10Hz. To mitigate the shortcut learning issue identified in \cite{PlanTF}, only the current state of the AV was used. Both the Scene Encoder and CMD consist of four stacked layers. In CMD, the first layer employs a single feed-forward network (FFN) without a Mixture of Experts (MoE), while the remaining layers incorporate two shared expert decoders and 16 routed expert decoders. When integrating with post-processing, we followed the method suggested in \cite{stateTransformer}. The detailed configurations for our model are provided in the Supplementary Material.

\begin{table}[t]
    \centering
    \setlength{\tabcolsep}{5pt}
    \fontsize{8pt}{9pt}\selectfont
    \caption{Ablation study of CarPLAN. \\
    SSE: Scene-Specific Experts, SAR: Scene-Aware Router.}
    \begin{tabular}{cccccc}
        \toprule[1.2pt]
        \multirow{2}{*}{\textbf{DPE}} & \multicolumn{2}{c}{\textbf{CMD}} &
        \multirow{2}{*}{\textbf{NR}$\uparrow$} & \multirow{2}{*}{\textbf{Colli}$\uparrow$} & \multirow{2}{*}{\textbf{DAC}$\uparrow$} \\
        \cmidrule(r){2-3}
        & \textbf{SSE} & \textbf{SAR} & & & \\
        \midrule[0.4pt]
        & & & 75.5 & 88.5 & 96.3\\
        \checkmark & & & 77.0  & 90.3 &  97.6 \\
        & \checkmark & & 76.5  & 89.0  &  96.8 \\
        & \checkmark & \checkmark & 77.4 & 90.6 & 98.1 \\ 
        \rowcolor[gray]{0.90} \checkmark & \checkmark & \checkmark & \textbf{78.6} & \textbf{92.1} & \textbf{98.5} \\ 
        \bottomrule[1.2pt]
    \end{tabular}
    \label{ablation_main}
    \vspace{-8pt}
\end{table}

\subsection{Performance Comparison}
Table \ref{tab:closed_loop_main} presents the performance of CarPLAN on the nuPlan dataset \cite{nuplan}, evaluated using OLS, CLS-NR, and CLS-R metrics. CarPLAN establishes new state-of-the-art performance across all CLS benchmarks, outperforming the latest planners by significant margins. While achieving superior CLS performance, CarPLAN maintains competitive OLS scores without significant compromise, which shows that CarPLAN exhibits human-like driving behavior.
On the Val14 benchmark, CarPLAN achieves a CLS-NR score of 91.4 and a CLS-R score of 84.6, surpassing the previous best models, Diffusion-Planner \cite{Diffusion-Planner} and BeTopNet \cite{BeTopNet}, by 1.5 and 0.9 points, respectively. On the Test14-Hard benchmark, CarPLAN also outperforms its competitors, demonstrating its ability to handle complex and challenging driving scenarios. 

Table \ref{table:hybrid} presents the performance in hybrid scenarios, where learning-based planners are combined with rule-based post-processing \cite{PLUTO, gameformer, stateTransformer}. CarPLAN achieves state-of-the-art performance across most benchmarks, except for CLS-NR on the Test14-Random benchmark, where it ranks second, trailing the best-performing model by a margin of 0.2 points.

\subsection{Ablation Study}
We conducted an ablation study to assess the contributions of CarPLAN's core components. Evaluation was conducted on the Test14-Hard benchmark using the CLS-NR metric. Additionally, we employed Colli (collision avoidance) and DAC (drivable area compliance) as supplementary evaluation metrics.

\subsubsection{Contributions of Main Components}
Table \ref{ablation_main} presents the contribution of each component to overall performance. 
The first row represents the baseline model, which consists of only the scene encoder and a Transformer decoder to generate future trajectories for the AV.
Integrating DPE into the baseline model increases the CLS-NR score by $1.5$ points, demonstrating the effectiveness of incorporating scene awareness. Next, we add Scene-Specific Experts without DPE, using a simple multi-layer perceptron (MLP)-based router that does not employ its own self- or cross-attention. This modification improves performance by $1.0$ points over the baseline.
When we enable both Scene-Specific Experts and Scene-Aware Router only without DPE, the CLS-NR score gap becomes $1.9$ points compared to the baseline. Finally, incorporating both DPE and CMD together results in a total improvement of $3.1$ points over the baseline. In particular, the Colli and DAC metrics improve by $3.6$ and $2.2$ points over the baseline, indicating that the proposed ideas significantly enhance safety in planning.

\begin{table}[t]
    \centering
    \setlength{\tabcolsep}{5pt}
    \fontsize{8pt}{9pt}\selectfont
    \caption{Ablation study for DPE target.}
    \vspace{5pt}
    \begin{tabular}{cccc}
        \toprule[1.2pt]
        \textbf{Displacement Target} & \textbf{NR}$\uparrow$ & \textbf{Colli}$\uparrow$ & \textbf{DAC}$\uparrow$ \\
        \cmidrule(lr){1-1} \cmidrule(lr){2-4}
        No Target  & 77.4 & 90.6 & 98.1 \\
        Agent Only & 78.1 & 91.6 & 98.0 \\
        Map Only   & 77.9 & 91.2 & 98.3 \\
        \rowcolor[gray]{0.90} Agent + Map & \textbf{78.6} & \textbf{92.1} & \textbf{98.5} \\
        \bottomrule[1.2pt]
    \end{tabular}
    \label{ablation_DPE}
    \vspace{-8pt}
\end{table}

\subsubsection{Ablation Study for DPE Target}
We explored different configurations of DPE by predicting future displacement vectors for: (1) No Target, (2) agents only, (3) map components only, and (4) both. Table \ref{ablation_DPE} presents the performance for each case.
Incorporating displacement predictions for surrounding agents alone improves the CLS-NR score by $0.7$ points over the No Target setting, while including HD map elements alone results in an improvement by $0.5$ points. When displacement vectors for both agents and map components are predicted, as in our CarPLAN, we observe a $1.2$ point improvement in CLS-NR, along with $1.5$ and $0.4$ point gains in Colli and DAC, respectively. These results confirm the enhanced awareness of both dynamic agents and traffic structures in our CarPLAN.

\begin{table}[t]
    \centering
    \setlength{\tabcolsep}{5pt}
    \fontsize{8pt}{8pt}\selectfont
    \caption{Ablation study for CMD.}
    \vspace{5pt}
    \begin{tabular}{cccccc}
        \toprule[1.2pt]
        \multicolumn{3}{c}{\textbf{Expert}} & \multirow{2}{*}{\textbf{NR}$\uparrow$} & \multirow{2}{*}{\textbf{Colli}$\uparrow$} & \multirow{2}{*}{\textbf{DAC}$\uparrow$} \\
        \cmidrule(lr){1-3}
        Routed & TopK & Shared \\
        \midrule[0.4pt]
        16 & 2 & - & 77.3  & 90.5 &  98.0  \\ 
        16 & 4 & - & 77.5  & 90.7 &  98.1  \\ 
        \rowcolor[gray]{0.90} 16 & 2 & \checkmark & \textbf{78.6}  & \textbf{92.1} &  \textbf{98.5}  \\ 
        16 & - & \checkmark & 77.9  & 91.2 &  98.3  \\
        16 & 1 & \checkmark & 77.2 & 90.5 & 97.7   \\ 
        16 & 4 & \checkmark & 78.1 & 91.5 & 98.3   \\
        8 & 2 & \checkmark & 77.7 & 91.1 & 98.2 \\
        32 & 2 & \checkmark & 77.5 & 90.8 & 98.1 \\
        \bottomrule[1.2pt]
    \end{tabular}
    \label{ablation_CMD}
    \vspace{-8pt}
\end{table}

\subsubsection{Effect of Expert Routing and Shared Experts}
Table \ref{ablation_CMD} analyzes the impact of the number of routed experts, Top-$K$
selection, and the inclusion of two additional shared experts on performance. The results show that the best performance is achieved when Top-2 experts are selected from 16 routed experts, with two shared experts enabled. Increasing or decreasing the number of experts to 32 or 8 does not lead to any performance improvement. Notably, the presence of two shared experts provides a significant performance boost. The shared experts are responsible for capturing general and context-invariant driving patterns, while the routed experts focus on scenario-specific behaviors. This complementary design allows the model to maintain globally consistent driving representations while adapting to diverse contexts, leading to improved robustness in planning.

\begin{table}[t]
    \centering
    \setlength{\tabcolsep}{6pt}
    \renewcommand{\arraystretch}{1.2}
    \fontsize{8pt}{9pt}\selectfont
    \caption{Inference Efficiency Analysis}
    \label{tab:ablation_efficiency}

    \begin{tabular}{l c c c}
        \toprule
        \textbf{Method} & \textbf{CLS-NR}$\uparrow$ & \textbf{FPS}$\uparrow$ & \textbf{FLOPs (G)}$\downarrow$ \\
        \midrule
        Baseline                & 75.5 & \textbf{17.74} & \textbf{0.94} \\
        Baseline + DPE          & 77.0 & \textbf{17.74} & \textbf{0.94} \\
        Baseline + CMD          & 77.4 & 14.94 & 1.08 \\
        \rowcolor[gray]{0.90}\textbf{CarPLAN} & \textbf{78.6} & 14.94 & 1.08 \\
        \bottomrule
    \end{tabular}
\end{table}


\begin{figure*}[!t]
    \centering
    \begin{minipage}{0.49\textwidth}
        \centering
        \includegraphics[width=\linewidth]{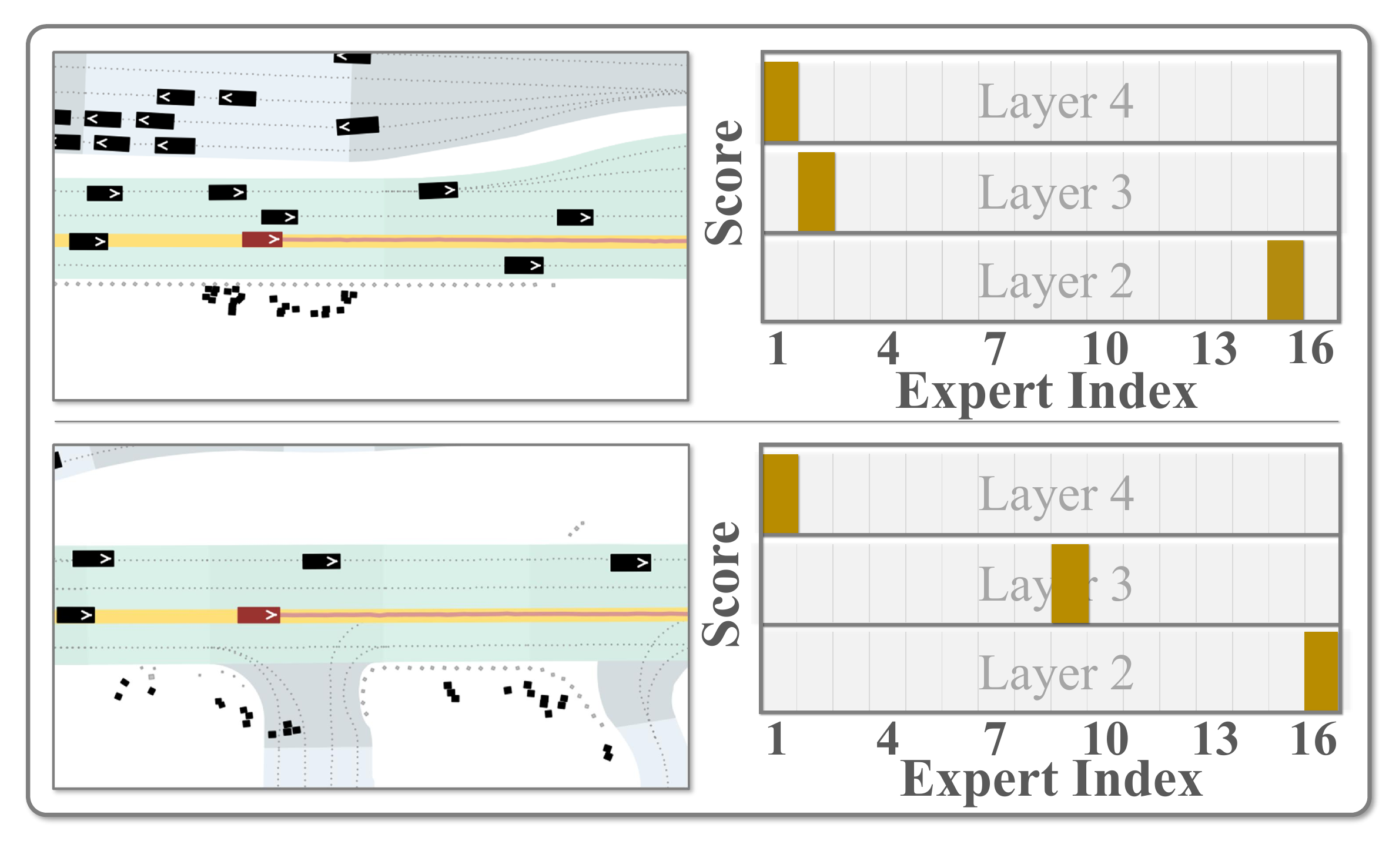}
        \vspace{-1mm}
        {\small (a)}
    \end{minipage}
    \hfill
    \begin{minipage}{0.49\textwidth}
        \centering
        \includegraphics[width=\linewidth]{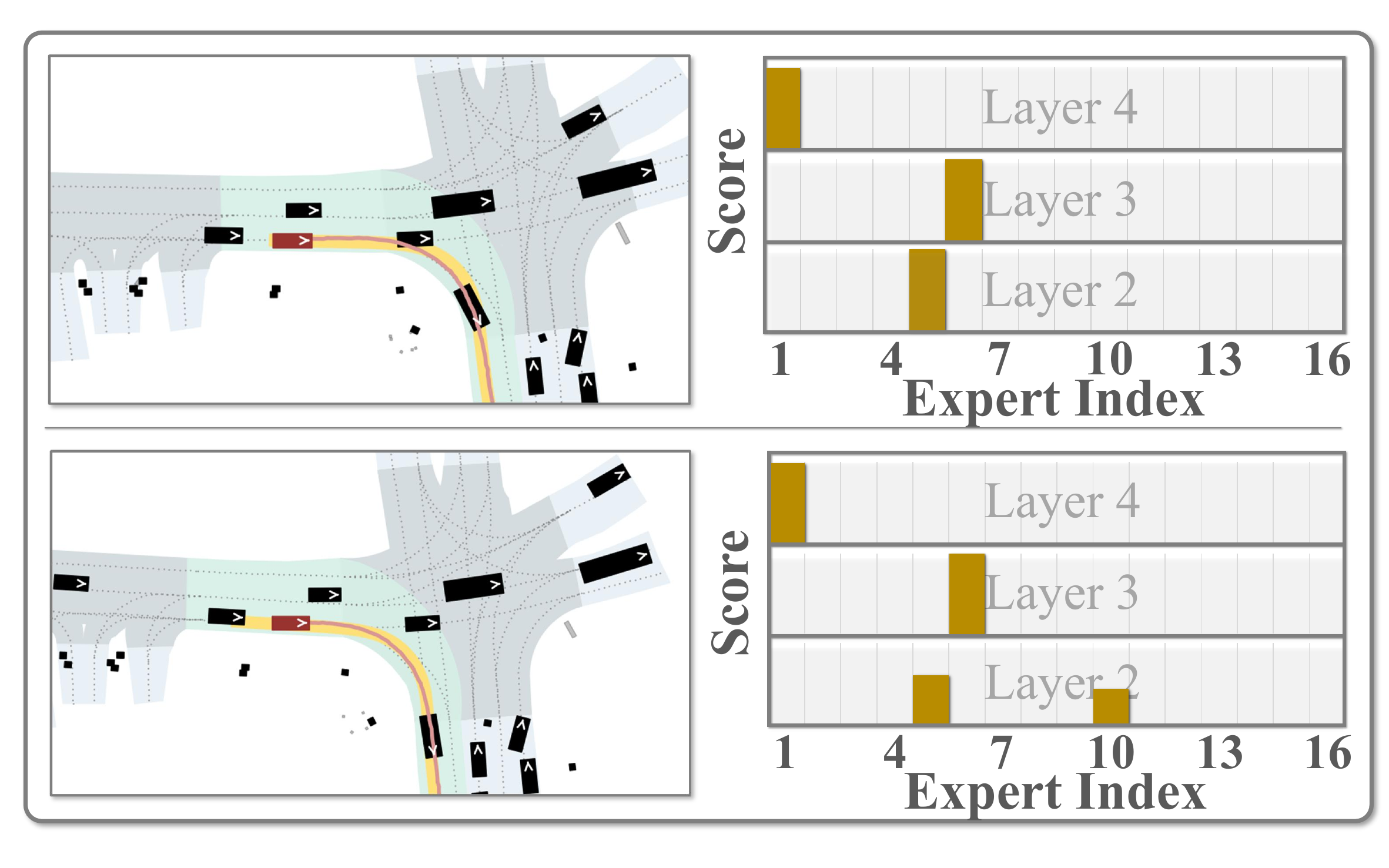}
        \vspace{-1mm}
        {\small (b)}
    \end{minipage}

    \caption{\textbf{Visualization of expert selection scores across layers.} 
   The dark red vehicle represents the AV, while the yellow and red lines indicate the ground truth (GT) and the predicted future trajectory with the highest score, respectively. The softmax scores are displayed for layers 2, 3, and 4.
(a) In two distinct straight-driving scenarios, differences in the distribution of surrounding agents lead to distinct expert selections.
(b) In similar driving scenarios, expert selection remains mostly consistent.}
    \label{Qualitative3_fig4}
\end{figure*}

\subsubsection{Inference Efficiency Analysis}
Table \ref{tab:ablation_efficiency} presents the efficiency analysis of CarPLAN under different module configurations. 
In our design, the Displacement-Aware Predictive Encoder (DPE) is used only during training, allowing it to be excluded during inference.
As a result, the inference-time computational cost mainly stems from the MoE-based decoder architecture. 
While the proposed method introduces a modest increase in latency and FLOPs, it still achieves real-time inference at approximately 15 FPS. 
More importantly, this overhead is accompanied by substantial performance gains, indicating a favorable efficiency–performance trade-off.
It is also worth noting that the current implementation does not apply any system-level optimization of the MoE structure.  Inference efficiency can be further improved through standard techniques such as expert pruning or parallel execution (e.g., CUDA streams).

\begin{table}[t]
    \centering
    \caption{Ablation study on the Waymax benchmark.}
    \label{tab:waymax}
    \setlength{\tabcolsep}{6pt}
    \renewcommand{\arraystretch}{1.2}
    \begin{tabular}{lcccc}
        \toprule
        \textbf{Method} & \textbf{AR}$\uparrow$ & \textbf{OR}$\downarrow$ & \textbf{CR}$\downarrow$ & \textbf{PR}$\uparrow$ \\
        \midrule
        Baseline & 87.52 & 1.77 & 2.61 & 95.34 \\
        Baseline + DPE & 88.21 & 1.52 & 2.47 & 95.89 \\
        Baseline + CMD & 88.56 & 1.60 & 2.51 & 96.32  \\
        \rowcolor[gray]{0.90}
        \textbf{CarPLAN} & \textbf{89.87} & \textbf{1.51} & \textbf{2.43} & \textbf{97.01} \\ 
        \bottomrule
    \end{tabular}
\end{table}

\subsubsection{Ablation Study on the Waymax Benchmark}
On the Waymax benchmark, Table \ref{tab:waymax} presents experimental results evaluating the effect of the proposed modules by integrating them into a baseline planner \cite{PLUTO}.
In this setting, the proposed DPE and CMD are incrementally incorporated into the encoder and decoder of the baseline model.
The results show that integrating either DPE or CMD individually yields consistent performance improvements, while their combined use achieves the highest performance across all evaluation metrics.
These findings indicate that displacement-aware relational encoding and context-adaptive decoding effectively contribute to performance improvements, and further demonstrate that the core design of CarPLAN generalizes well across different benchmarks.

\begin{figure*}[!t]
    \centering
    \begin{minipage}{0.49\textwidth}
        \centering
        \includegraphics[width=\linewidth]{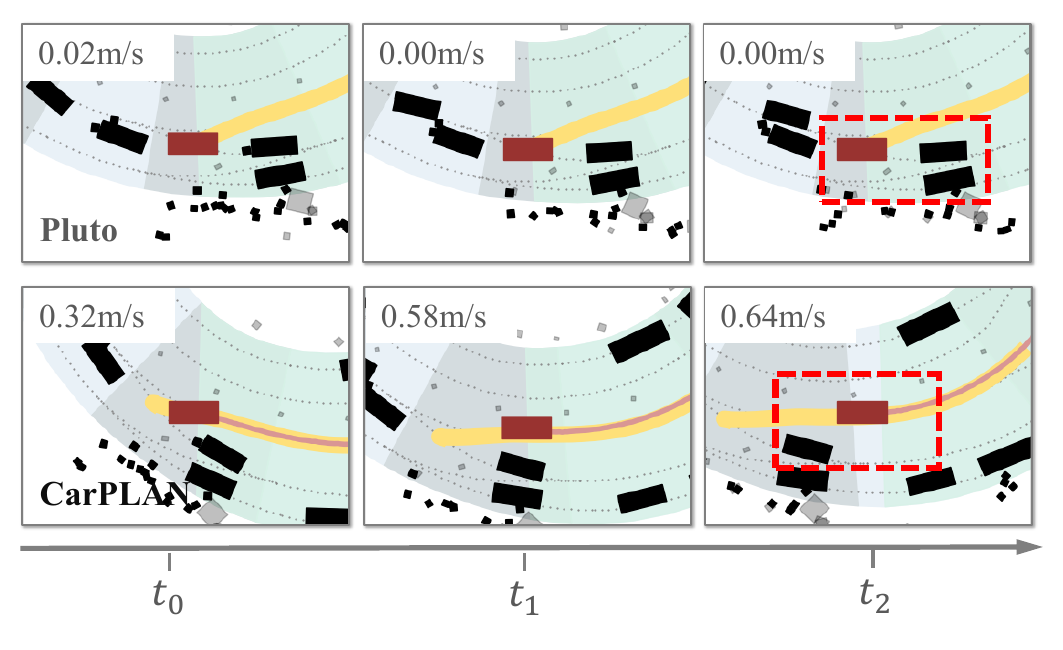}
        \vspace{-1mm}
        {\small (a)}
    \end{minipage}
    \hfill
    \begin{minipage}{0.49\textwidth}
        \centering
        \includegraphics[width=\linewidth]{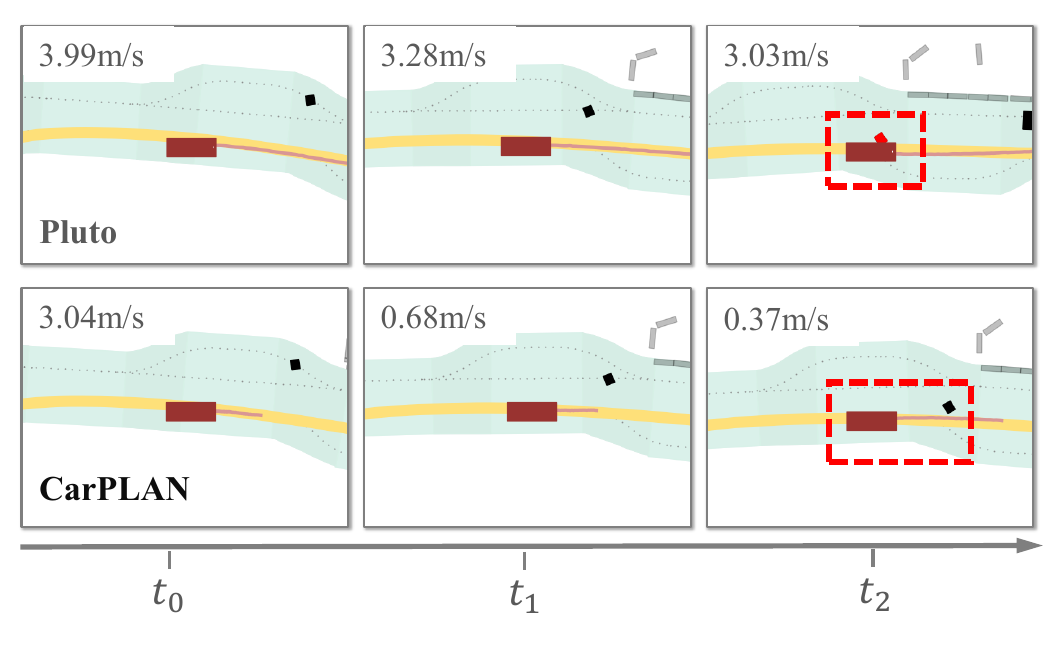}
        \vspace{-1mm}
        {\small (b)}
    \end{minipage}

    \caption{\textbf{Qualitative results on closed-loop simulations in nuPlan benchmark.} The yellow trajectory represents the recorded actual trajectory of the vehicle, while the red trajectory indicates the model's predicted trajectory with the highest probability at each timestep. The dark red vehicle represents the AV, and black vehicles or pedestrians that turn red signify a collision occurrence. Red dashed boxes highlight critical events observed during the simulation.}
    \label{Qualitative1_fig}
\end{figure*}


\begin{figure*}[!t]
    \centering
    \begin{minipage}{0.49\textwidth}
        \centering
        \includegraphics[width=\linewidth]{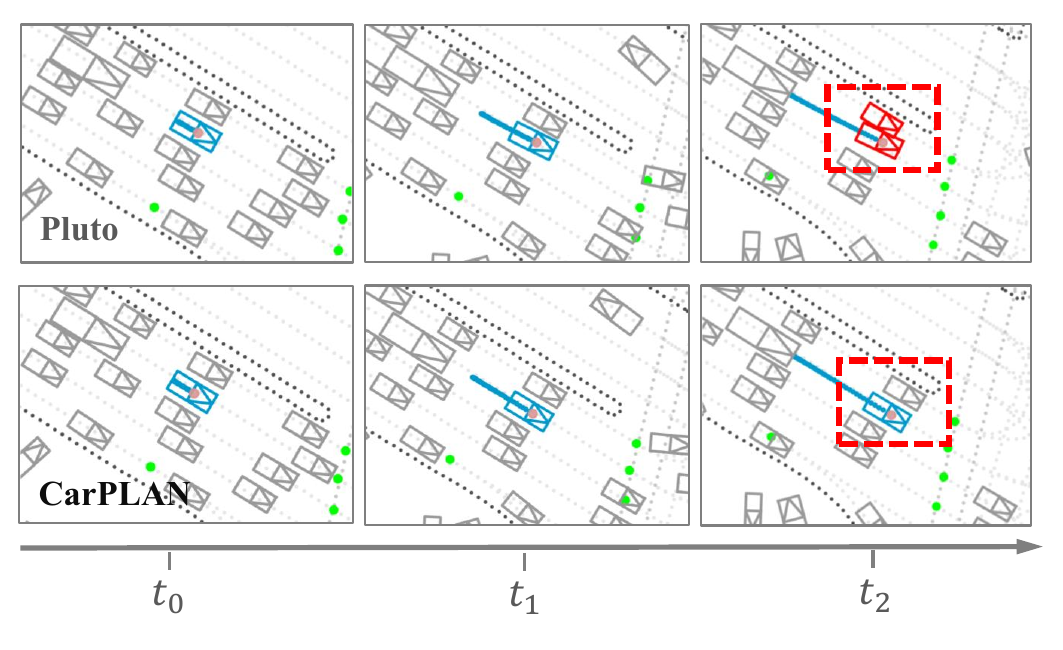}
        \vspace{-1mm}
        {\small (a)}
    \end{minipage}
    \hfill
    \begin{minipage}{0.49\textwidth}
        \centering
        \includegraphics[width=\linewidth]{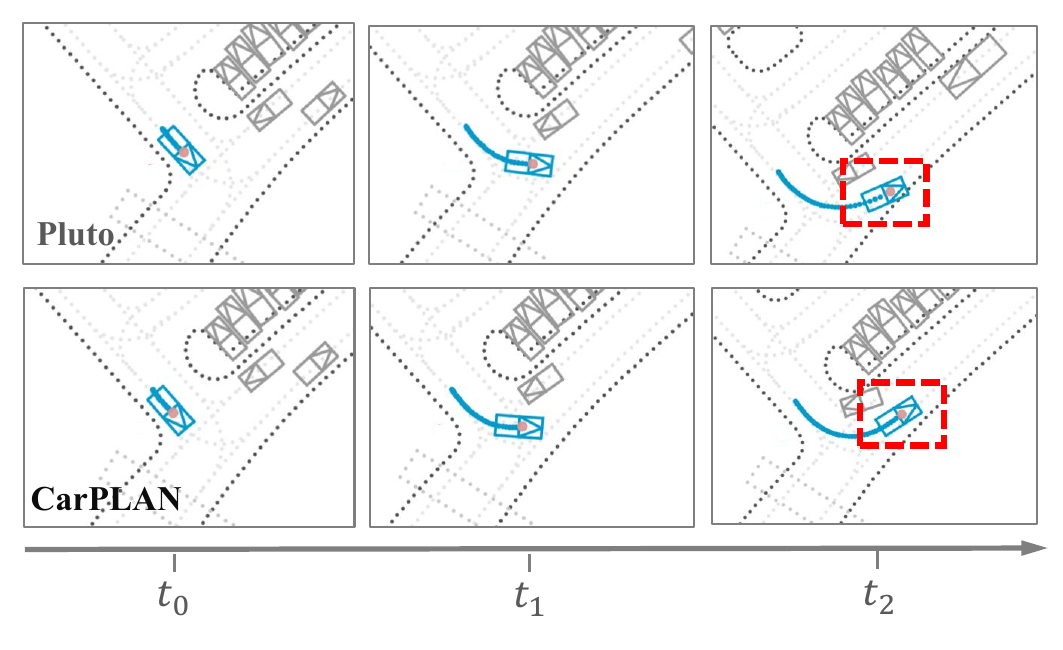}
        \vspace{-1mm}
        {\small (b)}
    \end{minipage}

    \caption{\textbf{Qualitative results on closed-loop simulations in Waymax benchmark.} The blue dots represent the trajectory executed during the closed-loop simulation. The blue bounding box denotes the AV, while gray bounding boxes indicate surrounding vehicles and boxes that turn red indicate a collision event during simulation. Green circles correspond to traffic lights in the green phase. Dark gray dots illustrate the road layout, and light gray dots represent centerlines. Red dashed boxes highlight critical events observed during the simulation.}
    \label{Qualitative2_fig}
\end{figure*}

\subsection{Qualitative Analysis}
\subsubsection{Visualization of Activated Expert's Score in CMD} 
To illustrate the expert selection behavior in CMD, Figure \ref{Qualitative3_fig4} visualizes the expert scores obtained at each layer across different driving scenes.
First, we observe that expert selection varies across CMD layers, indicating that the model dynamically adapts expert selection at different layers.
In Figure \ref{Qualitative3_fig4} (a), we examine two cases of straight-driving scenarios with different distributions of surrounding vehicles. The proposed Scene-Aware Router recognizes these as distinct situations and optimizes expert selection accordingly, producing different expert combinations.
Conversely, in Figure \ref{Qualitative3_fig4} (b), the two cases exhibit similar road topology and distributions of surrounding agents. In this scenario, the Scene-Aware Router selects a similar expert combination.

\subsubsection{Qualitative Results on closed-loop simulation} 
Figure \ref{Qualitative1_fig} presents a qualitative evaluation of the proposed CarPLAN, comparing its planning results with those of PLUTO \cite{PLUTO} in closed-loop simulations using nuPlan \cite{nuplan}. In Figure \ref{Qualitative1_fig} (a), PLUTO generates a conservative trajectory in a complex scene with dense traffic and pedestrians, whereas CarPLAN produces a more flexible trajectory appropriate for the current situation, enabling more context-adaptive decisions. Figure \ref{Qualitative1_fig} (b) demonstrates CarPLAN's ability to recognize a pedestrian crossing the road and appropriately reduce speed, highlighting its effectiveness in enhancing safety in dynamic environments.
Figure \ref{Qualitative2_fig} presents qualitative results from closed-loop simulations in Waymax \cite{waymax}, comparing CarPLAN with PLUTO. In Figure \ref{Qualitative2_fig} (a), PLUTO fails to maintain a safe distance from a nearby vehicle and results in a collision, whereas CarPLAN preserves an appropriate safety margin and avoids the collision. Figure \ref{Qualitative2_fig} (b) shows that PLUTO struggles to maintain sufficient clearance from the road boundary and deviates from the drivable area, while CarPLAN remains within the drivable region.

\section{Conclusion}
\label{sec:conclusion}

In this paper, we introduced CarPLAN, an IL-based planning method designed to more accurately capture human driving behavior. CarPLAN first integrates DPE, which models human-like behavior by maintaining relative spacing with surrounding agents and map components. DPE is trained to predict the future displacement vectors between the AV and scene elements, enabling the model to account for relative spacing in the planning process.
To handle complex and dynamic driving scenes, CarPLAN employs CMD, which facilitates context-adaptive decoding. CMD utilizes multiple expert decoders that are dynamically selected based on scene structure, allowing the model to effectively adapt to diverse driving conditions.
Our experiments on the nuPlan benchmark demonstrate that by integrating these components, CarPLAN consistently outperforms state-of-the-art learning-based planning methods in a variety of closed-loop simulations. Moreover, results on the Waymax benchmark indicate that CarPLAN generalizes well across different benchmarks.

Our study assumes the availability of perfect perception, relying on accurate object states and HD map information provided by external perception modules. As a future direction, we aim to extend our framework to an end-to-end autonomous driving system, where CarPLAN is jointly trained with the perception module. In addition, we plan to extend CarPLAN toward an action-conditioned world model that predicts how future spatial relationships evolve conditioned on the ego vehicle’s actions, rather than assuming a single deterministic displacement.


\bibliographystyle{IEEEtran}
\bibliography{IEEEtran}

\end{document}